\documentclass[letterpaper, 10 pt, conference]{ieeeconf}  

\IEEEoverridecommandlockouts                              

\usepackage{graphics} 
\usepackage{epsfig} 
\usepackage{mathptmx} 
\usepackage{times} 
\usepackage{amsmath} 
\usepackage{amssymb}  
\usepackage{caption}
\usepackage{subcaption}
\usepackage{soul}

\usepackage{xcolor}
\definecolor{mycolor1}{rgb}{1.00000,1.00000,0.00000}%
\definecolor{mycolor2}{rgb}{1.00000,0.00000,1.00000}%
\definecolor{mycolor3}{rgb}{0.00000,1.00000,1.00000}%

\usepackage{pgfplots}
\pgfplotsset{compat=newest}
\usetikzlibrary{plotmarks}
\usetikzlibrary{arrows.meta}
\usepgfplotslibrary{patchplots}
\usepackage{grffile}

\setstcolor{red}

\title{\LARGE \bf
Thruster-assisted Center Manifold Shaping in Bipedal Legged Locomotion
}

\author{Arthur C. B. de Oliveira$^{1}$ and Alireza Ramezani$^{2}$
\thanks{$^{1}$Arthur C. B. de Oliveira is a PhD student in Electrical and Computer Engineering, Northeastern University
        Boston, MA 02115, USA
        {\tt\small castello.a@northeastern.edu}}%
\thanks{$^{2}$Alireza Ramezani is a Faculty of Electrical and Computer Engineering, Northeastern University,
        Boston, MA 02115, USA
        {\tt\small a.ramezani@northeastern.edu}}%
}

\begin{document}

\maketitle
\thispagestyle{empty}
\pagestyle{empty}

\begin{abstract}
This work tries to contribute to the design of legged robots with capabilities boosted through thruster-assisted locomotion. Our long-term goal is the development of robots capable of negotiating unstructured environments, including land and air, by leveraging legs and thrusters collaboratively. These robots could be used in a broad number of applications including search and rescue operations, space exploration, automated package handling in residential spaces and digital agriculture, to name a few. In all of these examples, the unique capability of thruster-assisted mobility greatly broadens the locomotion designs possibilities for these systems. In an effort to demonstrate thrusters effectiveness in the robustification and efficiency of bipedal locomotion gaits, this work explores their effects on the gait limit cycles and proposes new design paradigms based on shaping these center manifolds with strong foliations. Unilateral contact force feasibility conditions are resolved in an optimal control scheme. 

\end{abstract}

\section{INTRODUCTION}

This work tries to contribute to the design of legged robots with capabilities boosted through thruster-assisted locomotion and capable of negotiating unstructured environments, including land and air, by leveraging their legs and thrusters collaboratively. These robots could be used in a broad number of applications including search and rescue operations, space exploration, automated package handling in residential spaces and digital agriculture, to name a few. In all of these examples, the unique capability of thruster-assisted mobility greatly broadens the locomotion designs possibilities for these systems. 

For instance, in Search and Rescue (SAR) operations in the aftermath of unique incidents new catastrophic events might follow. A hurricane may produce flooding or the collapse of structures due to wind damage, a landslide may dam a river and create a flood. In these scenarios, these robots can leverage their hybrid mobility and adapt to the search mission. These robots, which have been remarkably overlooked in SAR operations, can deliver important strategic situational awareness involving aerial survey, and reconnaissance through multi-purpose scans of the area with the suite of sensors integrated in their designs. Airborne structural inspection of building in harsh atmospheric conditions is not possible and aerial mobility is not practical inside collapsed buildings, however legged systems can leverage their legged mobility in the form of crawling or walking and inspect inside these structures.

\begin{figure}[t]
\vspace{0.1in}
    \centering
    \includegraphics[width=0.6\linewidth]{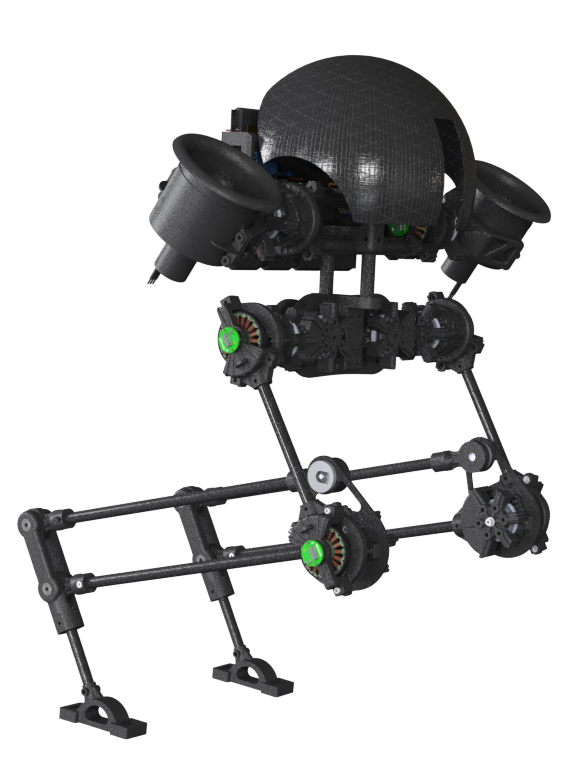}
    \caption{NU's hybrid legged-aerial robot, \textit{Harpy}}
    \label{fig:Leo}
\vspace{-0.1in}
\end{figure}

In the aftermath of Hurricane Katrina, in 2005, that set the stage for drone deployments in disaster-affected region, regulations promulgated by the Federal Aviation Administration (FAA) posed severe limitations in practicality and usefulness of aerial drones in inflicted regions. According to AFF, extreme care is needed when flying near people, because operators tend to lose depth perception and may get too close to objects and people. In addition, some platforms or payloads may not be able to maneuver safely for this mission type. For instance, something hanging off a small drone changes the dynamics of the vehicle, creating a pendulum effect which may cause unpredictable behaviours. The idea of integrating other modes of locomotion with the safety of legged systems to accelerate and facilitate SAR operations, using them in the delivery of food and medical supplies or even just in the search for survivals, can transform  and potentially increase their effectiveness. In Fig. \ref{fig:Leo} one example of such robot is presented.

Such hybrid mobility systems have been considered to a limited level in the past, approaches involving articulated tracks and wheels such as \cite{shen2009design, kim2014wheel} leverage the agility of wheeled platforms and versatility of legged systems, but legged-wheeled systems rarely take advantage of the integration between the two different modalities of locomotion. Similarly, in flying-crawling designs such as in \cite{daler2013flying} the two modes of locomotion are often treated as independent tools, rarely being integrated. 



Other than their strong and impactful applications in SAR operations, these hybrid systems are interesting modeling and control problems. From a control design standpoint, the inspection of these robots can help expand the knowledge of reduced-order models (ROMs) that involve fluidic force interactions applied to aerial and aquatic locomotion systems. Earlier legged locomotion works \cite{ramezani_performance_2014,ramezani_atrias_2012,park_finite-state_2013,park_switching_2012,buss_preliminary_2014,dangol_performance_2020,dangol_thruster-assisted_2020,dangol_towards_2020} have convinced us that reduced-order systems, in an exercise of creative neglect, can simplify or ignore dynamics redundancy. They have been invaluable in uncovering basic legged dynamical structures as they are described by the smallest number of variables and parameters required for the exhibition of a behavior of interest, which also can be hypothesized as attracting invariant submanifolds in the state-space of the system.  

In a legged robot, the restricted dynamics on these embedded submanifolds take a form prescribed by the supervisory controller \cite{westervelt2018feedback}. For instance, the spring-loaded-inverted-pendulum (SLIP) model introduced by \cite{blickhan1993similarity,cavagna1977mechanical,altendorfer2001evidence} is a classical, celebrated ROM that describes the center of mass behavior of diverse legged animals. From a dynamical systems standpoint, this collapse of dimension in state-space would follow from the existence of an inertial manifold with a strong stable foliation \cite{guckenheimer1984nonlinear,constantin2012integral}. 

We note that while mathematical ROMs of legged robots of varying size and complexity are relatively well developed, such ROMs of airborne or fluidic-based locomotion remain largely open due to the complex fluid-structure interactions involved in their locomotion. Study of these models can potentially yield significantly robust legged systems not reported before. 
Robust legged locomotion has been studied extensively in the past and Boston Dynamic's BigDog \cite{raibert2008bigdog} and Raibert's hopping robots \cite{raibert1984experiments} are arguably amongst the most successful examples of legged robots, as they can hop or trot robustly even in the presence of significant unplanned disturbances. A large number of humanoid robots have also been introduced. Honda's ASIMO \cite{hirose2007honda} and Samsung's Mahru III \cite{kwon2007biped} are capable of walking, running, dancing and going up and down stairs. Despite these accomplishments, all of these systems are prone to falling over. Even humans, known for agile and robust gaits, whose performance easily outperform that of today's bipedal robot cannot recover from severe pushes or slippage on icy surfaces. A distributed array of thrusters can significantly enhance the robustness of these systems. The thrusters add to the array of control inputs in the system (i.e., adds to redundancy and might lead to overactuation) which can be beneficial from a practical standpoint and challenging from a feedback design standpoint.

In an effort to demonstrate thrusters effectiveness in the robustification and efficiency of the gaits, this work explores their effects on the gait limit cycles and proposes new design paradigms based on shaping these center manifolds with strong foliations. Unilateral contact force feasibility conditions are resolved in an optimal control scheme. 

In section \ref{sec:ThrsterZDyn}, we briefly re-visit the derivation of zero dynamics equations. In our derivations, the thruster roles are considered and as such it is shown how the restricted dynamics are affected by them. Then, results from \cite{manchester2014transverse} are applied to assume stable limit cycles for the thruster-augmented dynamics. We leverage the fact that the thrusters are usually much quicker than the walking dynamics and study how constant thruster forces change the resulting limit cycle and how instantaneous changes in that value during a step can be used to change the shape of the limit cycle without changing the pre- and post-impact states.

In section \ref{sec:ContForc}, we explore how the thrusters affect the contact forces both during the continuous phase and the impact. The ground contact forces are affine-in-thruster-action and are approximated by polynomial functions of the gait-timing variable and are enforced as constraints. 

\section{Thruster and Zero Dynamics}
\label{sec:ThrsterZDyn}

\begin{figure*}[ht]     
\vspace{0.1in}
    \begin{subfigure}{.49\textwidth}
        \centering
        \includegraphics[width=1\linewidth]{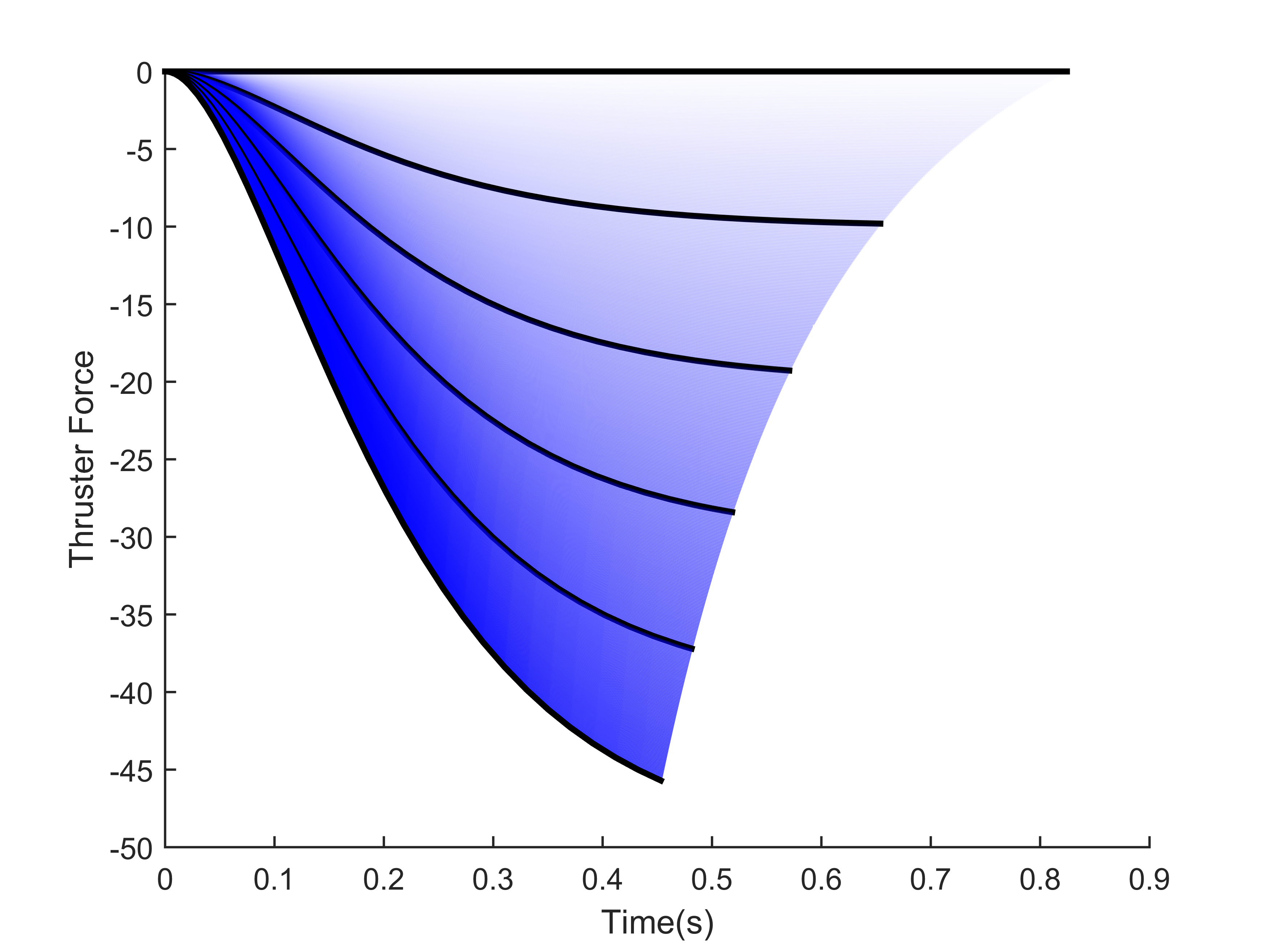}
        \caption{\centering Slowly changing thrusters force, steady state value varying from 0 (light color) to -50N (dark color), black lines are the increments of -10N.}
        \label{fig:SlowTHR_Ft}
    \end{subfigure}
    \begin{subfigure}{.49\textwidth}
        \centering
        \includegraphics[width=1\linewidth]{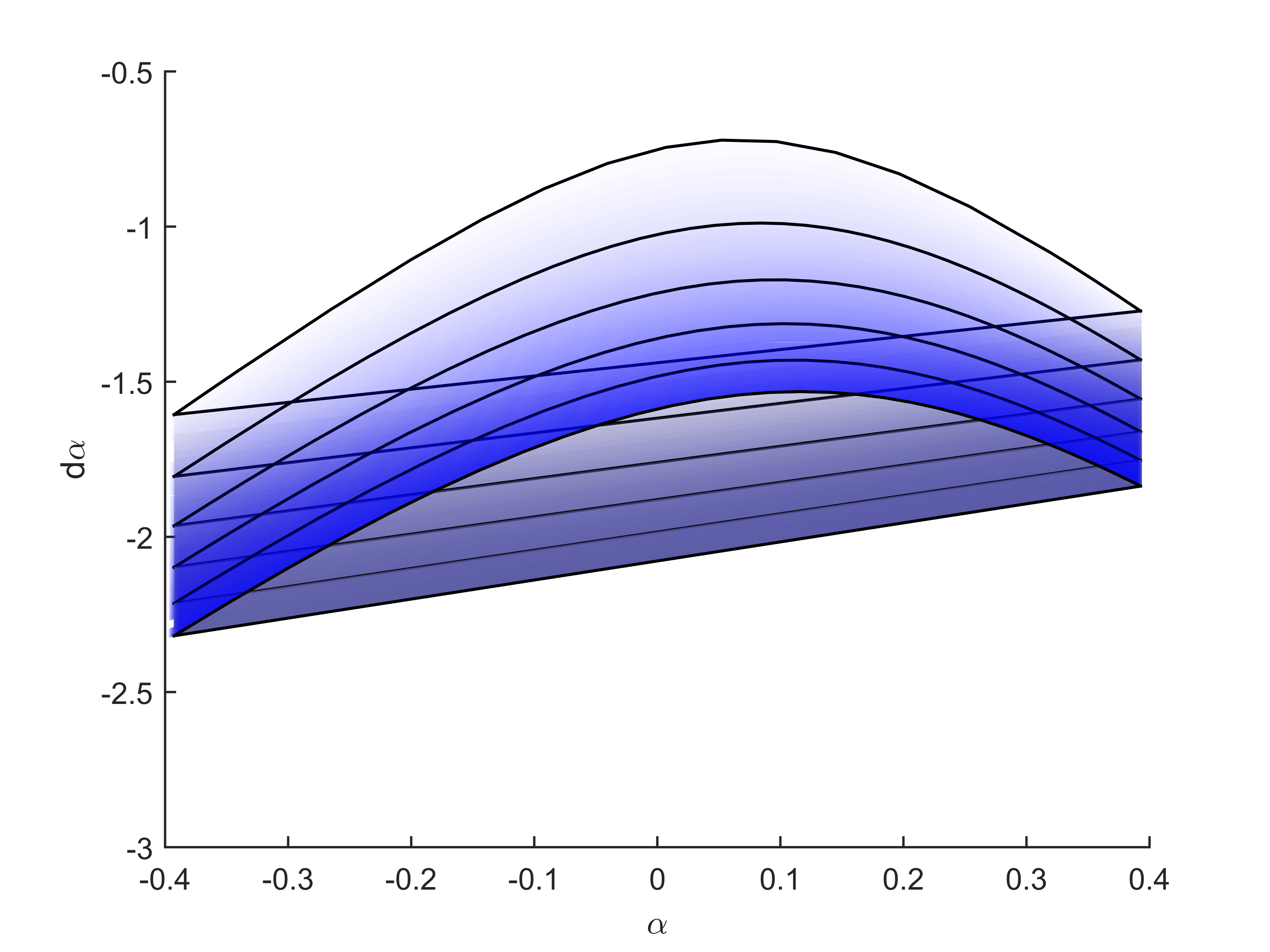}
        \caption{\centering Limit Cycles for slowly changing thruster force}
        \label{fig:SlowTHR_LC}
    \end{subfigure}
    \caption{Simulation results for thrusters modeled as second-order linear systems with slow dynamics.}
    \label{fig:SlowThruster}
    \vspace{-0.1in}
\end{figure*}

In this section, we compute how the thrusters affect the zero dynamics. The equations of motion of the robot can be obtained from the following Euler-Lagrange equation
%
%
%
\begin{equation}
    \label{eq:NonLinSys}
    \underbrace{\begin{bmatrix}
        D_{bb} & D_{bN} \\ D_{Nb} & D_{NN}
    \end{bmatrix}}_{D(q_b)}
    \begin{bmatrix}
        \ddot{q}_b \\ \ddot{q}_N
    \end{bmatrix} + 
    \underbrace{\begin{bmatrix}
        \Omega_b(q,\dot{q}) \\ \Omega_N(q,\dot{q})
    \end{bmatrix}}_{C(q,\dot{q})+G(q)} = 
    \underbrace{\begin{bmatrix}
        I & b_b \\ 0 & b_N
    \end{bmatrix}}_{[B_\tau \| B_F]}\begin{bmatrix}
        u \\ F_T
    \end{bmatrix},
\end{equation}

\noindent where $D$, $C$ and $G$ are the \textit{inertia matrix}, the \textit{coriolis and centripetal forces vector} and the \textit{gravitational vector} of the pinned system, respectively. The vector $q = [q_b ; q_N]$ represents the generalized configuration variables of the robot, where $q_b$ and $q_N$ are the vector of relative and absolute coordinates, respectively. Through the virtual work theorem $B_\tau$ can be obtained for the actuated joints torques $u$, and has the form $B_\tau = [I ; 0]$ if the relative coordinates $q_b$ are chosen as being each of the joints angles. Matrix $B_F = [b_b ; b_N]$ is also obtained for the thruster force $F_T = [F_T^x ; F_T^y]$ applied at a specific point of the robot through the virtual work theorem.

Choosing the feedback control law of the form%
\begin{equation}
    \label{eq:MPFL_Feedback}
    u = (D_{bb}-D_{bN}D_{NN}^{-1}D_{Nb})v+\Omega_b-D_{bN}D_{NN}^{-1}\Omega_N-b_bF_T,
\end{equation}
\noindent where $v$ is a new input to be designed for the system, the resulting closed loop system is presented in \eqref{eq:MPFL_SS}, where $\sigma_N = [D_{Nb} \ D_{NN}]q$ is the angular momentum about the stance foot (pivot point).
\begin{equation}
    \label{eq:MPFL_SS}
    \dot{x} = \begin{bmatrix}
        \dot{q}_b \\ \dot{q}_N \\ \ddot{q}_b \\ \dot{\sigma}_N
    \end{bmatrix} = \underbrace{\begin{bmatrix}
        \dot{q_b} \\ D_{NN}^{-1}\sigma_N-D_{NN}^{-1}D_{Nb}\dot{q}_b \\ 0 \\ b_NF_T-\frac{\partial V}{\partial q_N}
    \end{bmatrix}}_{f(x)}+\underbrace{\begin{bmatrix}
        0 \\ 0 \\ I \\ 0
    \end{bmatrix}}_{g(x)}v
\end{equation}
\noindent Taking a gait-timing variable $\alpha(x) = cq$ as a linear combination of the configuration variables so that $\alpha$ is strictly increasing or decreasing during a step, we can define an output function $y$ shown in \eqref{eq:output}, where $h_d(\alpha)$ are a set of $N-1$ $M$-th order bezier polynomials with parameter matrix $A$, that is, $a_{ij}$ is the $j$-th parameter of the $i$-th polynomial.
\begin{equation}
    \label{eq:output}
    y = q_b-h_d(\alpha), \qquad
    \dot{y} = \dot{q}_b-\frac{\partial h_d}{\partial \alpha}\dot{\alpha}
\end{equation}
The standard form for the full-dynamics is
\begin{equation}
    \label{eq:fullsys}
    \begin{bmatrix}
        \dot{y} \\ \dot\alpha \\ \ddot{y} \\ \dot\sigma_N
    \end{bmatrix} = \begin{bmatrix}
        \dot{y} \\ L_f\alpha(x) \\ L^2_fy(x)+L_gL_fy(x)v \\ L_f\sigma_N(x)
    \end{bmatrix},
\end{equation}
\noindent where $L_g(\cdot)$ and $L_f(\cdot)$ are the Lie derivatives on the vectorfields $f$ and $g$ respectively. $y\equiv0$ and $\dot{y}\equiv0$ are enforced through the feedback linearization control law 
\begin{equation}
    v = -L_gL_fy^{-1}(L^2_fy+K_d\dot{y}+K_py),
\end{equation}
\noindent with $K_d$, $K_p>0$. This controller yields the zero dynamics of the form of
\begin{equation}
    \label{eq:ZDyn}
    \begin{aligned}
    \dot{\alpha} &= (\tilde{D}_{NN}-\tilde{D}_{Nb}\tfrac{\partial h_d}{\partial \alpha})^{-1}\sigma_N = \kappa_1(\alpha)\sigma_N \\
    \dot{\sigma}_N &= -\tfrac{\partial V}{\partial \alpha}+b_NF_T = \kappa_2(\alpha)+b_NF_T,
    \end{aligned}
\end{equation}
\noindent where the matrix $\tilde{D}$ is the inertia matrix written in the coordinates $\tilde{q} = [q_b ; \alpha]= Hq$, where $H = [H_0 ; c]$ and $H_0 = [I_{N-1}, 0_{N-1 \times 1}]$ and we can define $\tilde{D} = (H^{-1})^\top DH^{-1}$.

\subsection{Thruster Dynamics and Existence of Limit Cycles}

Note that to obtain the zero dynamics in \eqref{eq:ZDyn}, no assumption is made about the thruster dynamics. Consider that for a given set of parameters $A$ the hybrid zero dynamics computed for $F_T\equiv 0$ has a stable limit cycle, then, \textit{Theorem 4} of \cite{manchester2014transverse} states that the continuous part of the zero dynamics necessarily has a Riemannian-like contraction metric that satisfies the transversal contractivity condition. Assuming that the dynamics $\dot{F}_T = g_F(F_T)$ that the thruster has is contractive, \textit{Theorem 5} of the same paper states that the cascade of the continous dynamics (transverse contractive) with the thrusters (contractive), as in \eqref{eq:cascade}, is transverse contractive and, therefore, has a stable limit cycle.
\begin{equation}
    \label{eq:cascade}
    \begin{gathered}
        \dot{F}_T = g_F(F_T) \\
        \dot{z} = \begin{bmatrix}
            \dot\alpha \\ \dot\sigma_N
        \end{bmatrix} = \begin{bmatrix}
            \kappa_1(\alpha)\sigma_N \\ \kappa_2(\alpha)+b_NF_T
        \end{bmatrix} = f_Z(z, F_T)
    \end{gathered}
\end{equation}

\begin{figure}[t]
\vspace{0.1in}
    \centering
    \includegraphics[width=0.7\linewidth]{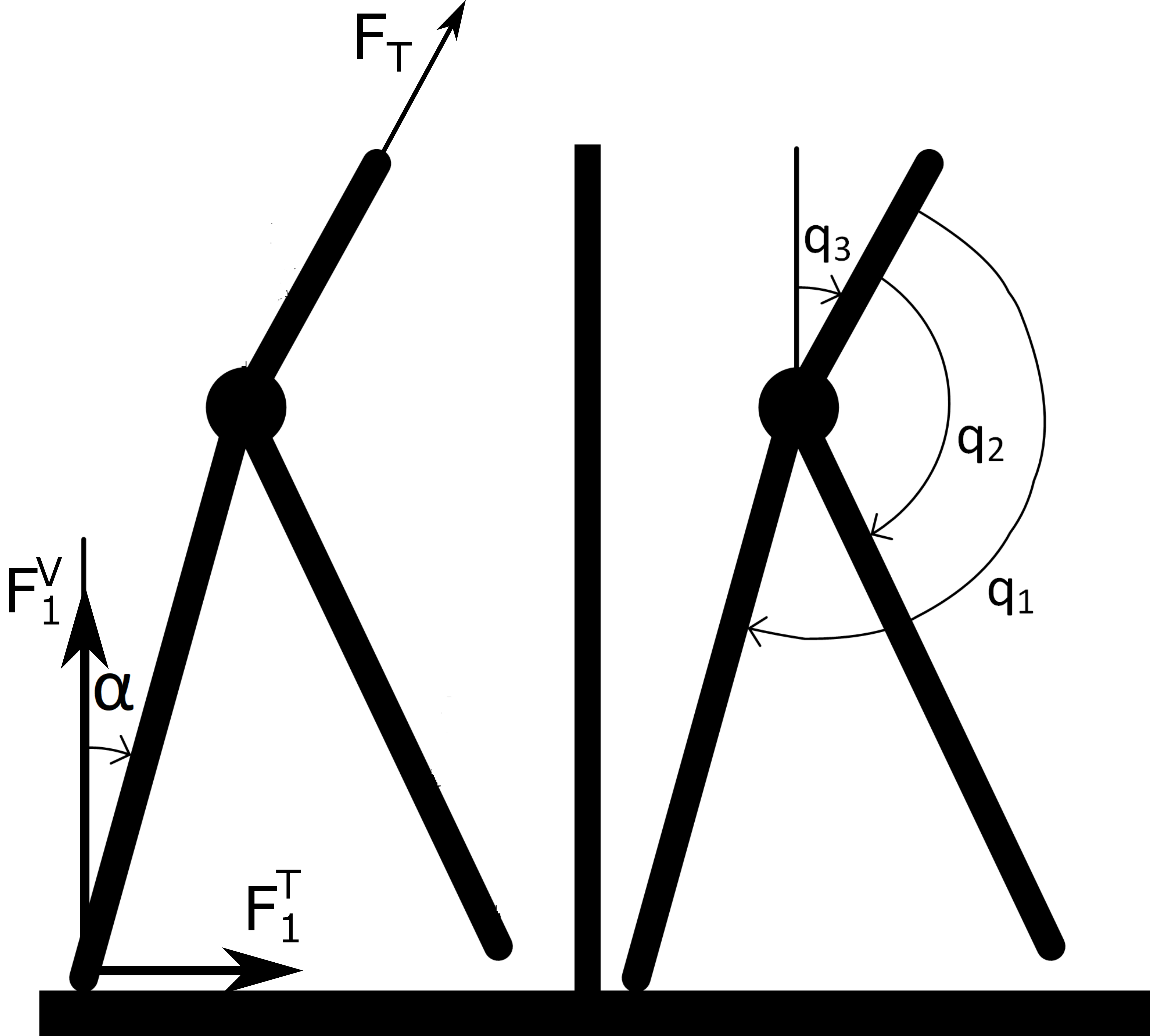}
    \caption{3-link model for a walking bipedal robot with gait timing variable $\alpha$. The stance foot position is assumed to be fixed to the ground.}
    \label{fig:3link_model}
    \vspace{-0.1in}
\end{figure}

To illustrate this, the simulation results for a 3-link robot, shown in Fig. \ref{fig:3link_model}, are presented for walking with slow thruster dynamics when compared to the unactuated dynamics. Figure \ref{fig:SlowTHR_Ft} shows the time evolution of the thruster action, which is assumed to be the response of a second-order system with the steady-state solution denoted by $F_{ss}$. Figure \ref{fig:SlowTHR_LC} illustrates the resulting limit cycles. While the simulation presented here are obtained from a simple model, note that the theoretical results are independent of the dimension of the robot and are valid for a 5-link or even more complex planar walkers assisted by thrusters.

While the thruster dynamics are considerably fast when compared to the internal dynamics for slow walking feats, faster and agile maneuvers such as running and jumping demand reasonably comparable dynamical descriptions of both sub-systems. As such, in addition to two-time-scale description of the thruster and legged dynamics, the dynamics in the form of $\dot{F}_T = f_{F_T}(t)$ are considered for the thrusters.  

The result reported in \cite{manchester2014transverse} assures the existence of stable limit cycles in the cascade systems composed of the internal and thruster dynamics. As a result, each dynamical model can be looked upon separately in a hierarchical design scheme. First, existing gait design approaches that assume stable supervisory controller and are widely used in legged locomotion can be applied to render the internal dynamics transverse contractive. Second, the thrusters are applied to adjust the resulting restricted dynamics on the zero-dynamics manifold. In the next section, we look at the results when considering the thrusters as instantaneously changing parameters of the zero dynamics. 

\subsection{Center Manifold Shaping Using Thrusters}

\begin{figure}[t]
\vspace{0.1in}
    \centering
    \includegraphics[width=1\linewidth]{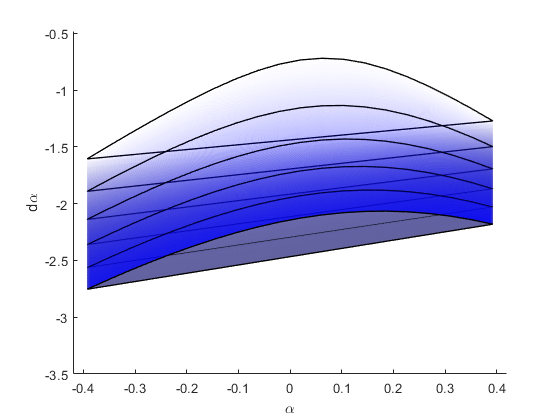}
    \caption{Limit cycles in a two-time-scale thruster-augmented dynamical model.}
    \label{fig:SingVal_THR}
    \vspace{-0.1in}
\end{figure}

In this section, we will apply the thruster action to modify the center manifolds (CM). The shape of these CMs are defined by the overall supervisory controller in the closed-loop system. However, in our thruster-assisted legged problem, the thrusters provide another option for such adjustments. In doing this, we will assume a two-time-scale problem wherein the thruster action is considered as a parameter with no dynamics. Applying the coordinate change for the zero dynamics $\zeta = \sigma_N^2 / 2$ and computing the partial derivative $\partial \zeta / \partial \alpha$ yields the solution for $\zeta(\alpha)$ presented in 
\begin{equation}
    \zeta(\alpha) = \zeta_i+\int_{\alpha_i}^{\alpha}  \left(  \frac{\kappa_2(\tau)}{\kappa_1(\tau)} + \frac{b_NF_T}{\kappa_1(\tau)} \right) d\tau,
    \label{eq:orig_integral}
\end{equation}
\noindent where $\zeta_i$ and $\alpha_i$ are the values of $\zeta$ and $\alpha$ at the beginning of the step. Notice that $\zeta_0(\alpha, \zeta_i) = \zeta_i+\int_{\alpha_i}^{\alpha} (\kappa_2(\tau) / \kappa_1(\tau)) d\tau$ is the nominal solution of the system, that is, when $F_T\equiv 0$. Defining $b_F(\alpha_j, \alpha_k) = \int_{\alpha_k}^{\alpha_j} (1/\kappa_1(\tau)) d\tau$, we can rewrite the solution of the zero dynamics as
%
%
%
\begin{equation}
    \label{eq:ZDSol}
    \zeta(\alpha) = \zeta_0(\alpha,\zeta_i) + b_F(\alpha, \alpha_i)b_NF_T.
\end{equation}
\noindent From Eq. \eqref{eq:ZDSol} one can see the direct effect of the thrusters as they linearly change the fixed-point of the Poincare function. In Fig. \ref{fig:SingVal_THR}, the limit cycles  when the thruster magnitude is adjusted incrementally from 0 to -50N are shown. Two adjustments are noticeable: 1) translation along y axis and 2) reduction in the area encircled by the limit cycle. While the first is a direct consequence from the change in the impact velocity caused by the extra energy added by the thruster, the second comes from the fact that the potential energy barrier of the gait is independent from the thruster action. That is, the gait needs to give in the same absolute amount of kinetic energy to surpass the potential energy barrier, independently from how much kinetic energy it has during the nominal gait, this makes such loss of velocity (represented in the graph by the point of minimum absolute velocity) less relevant in relation to the total energy of the system the more the thrusters increase the total energy of the system.

\begin{figure*}[ht]
\vspace{0.1in}
    \begin{minipage}{.5\textwidth}
        \centering
        \includegraphics[width=1\linewidth]{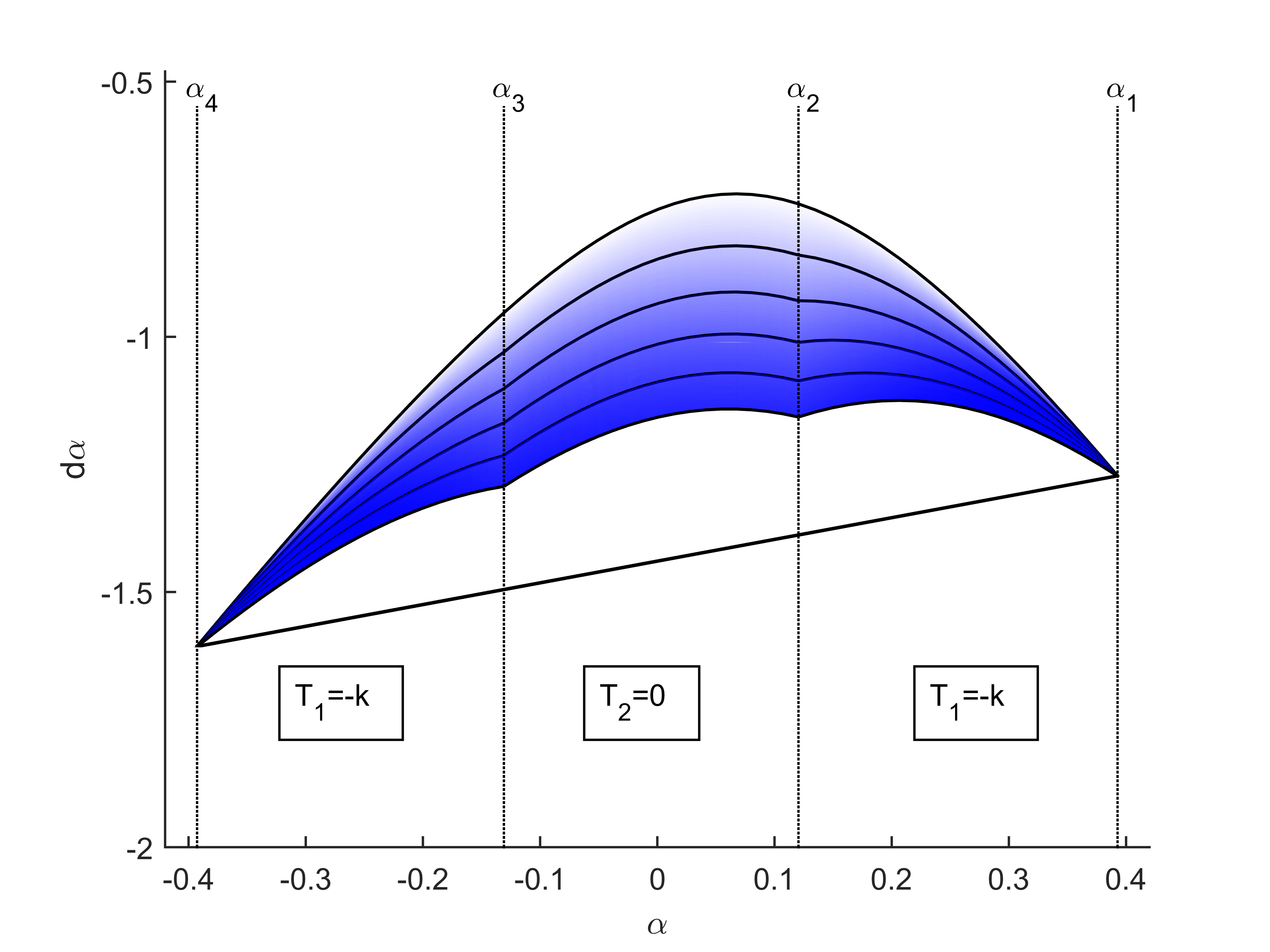}
        \caption{Limit Cycles for n=4, $T_j = k(j-2)$ and $k$ varying from 0 to 50 (black lines are the increments of 10 for k).}
        \label{fig:ThreeVal_THR}
    \end{minipage}
    \begin{minipage}{.5\textwidth}
        \centering
        \resizebox{1\textwidth}{!}{\input{figures/LimitCycles_Param_ThreeValues_60_wVF.tikz}}
        \caption{Change in the vector field for $k = 60$.}
        \label{fig:ThreeVal_VF}
    \end{minipage}
\vspace{-0.1in}
\end{figure*}

To shape the limit cycles, consider point-wise gait-timing variables  $\alpha_j = {\alpha_1,\dots,\alpha_n}$, where at the boundaries $\alpha_i=\alpha_1$ and $\alpha_n=\alpha_f$. The value of $b_NF_T$ at $\alpha_i$ is changed to some constant value $T_i\in\mathbb{R}$ instantaneously. This causes $b_N F_T / \kappa_1(\alpha)$ to be discontinuous at the boundaries yet it will remain continuous between the point-wise modifications of $T_i$ at $\alpha_i$.The new fixed-point of the Poincar\'e function $\zeta^*$ can be computed as
\begin{equation}
    \zeta^* = \zeta_0^*+\sum_{j=1}^{n-1}b_F(\alpha_{j+1}, \alpha_j)T_{j},
\end{equation}
\noindent which allows any desired value for the fixed-point $\zeta_d^* = \zeta_0^*+c$ through the constraint $\sum_{j=0}^{n-1}b_F(\alpha_{j+1}, \alpha_j)T_{j}-c=0$. Note that the sum of terms appears because the original integral in \eqref{eq:orig_integral} is broken between each consecutive pair of discontinuous points. This result can be interpreted in another way. That is, the gait limit cycles are adjustable by the thruster-assisted variations of the vector field restricted to each gait-timing variable envelope in the phase portrait. Figure \ref{fig:ThreeVal_THR} shows the variations in the limit cycles when the thruster magnitude is adjusted at 4 points along the gait cycle in increments from 0 to 50N. The points $\alpha_2$ and $\alpha_3$ were chosen so that $\zeta^* = \zeta_0^*$ holds. In Fig. \ref{fig:ThreeVal_VF}, the vector fields are illustrated. 

\section{Unilateral Contact Forces}
\label{sec:ContForc}

While the thrusters can be useful tools in shaping the gait limit cycles, special care must be taken when considering the contact force constraints, since the thruster could easily violate them. In this section, we take a better look at the contact forces during both the continuous and discrete dynamics of walking, and will show how thrusters can help. 

\subsection{Impact Force Constraints} 

Considering the gait feasibility hypotheses from \cite{westervelt2018feedback} for the impact, the resulting impact force $F_2$ can be obtained by integrating \eqref{eq:ImpMod} for the duration of the impact.
\begin{equation}
    \label{eq:ImpMod}
    D_u(q_u)\ddot{q}_u+\underbrace{C(q_u, \dot{q}_u)+G(q_u)}_{\Omega_u} = \begin{bmatrix}B_\tau & B_F \\ 0 & B_{Fu}\end{bmatrix}\begin{bmatrix} u \\ F_T \end{bmatrix} + \delta F_{ext}.
\end{equation}
\noindent In \eqref{eq:ImpMod}, $q_u = [q_b ; q_N ; x_1 ; y_1]$ is the generalized unpinned configuration vector, where $x_1$ and $y_1$ are the horizontal and vertical position of the stance foot, respectively. Also, $D_u$, $C_u$ and $G_u$ are the unpinned inertia and Coriolis matrices and the unpinned gravitational vector, respectively, and $B_{Fu} = [b_x ; b_y] = [\cos(\theta) ; \sin(\theta)]$, where $\theta$ is the angle between the horizontal axis and the thruster force. Terms $q_b$, $q_N$, $B_\tau$ and $B_F$ are defined as in  \eqref{eq:NonLinSys}. The impulsive and external force that acts at the swing leg end is denoted by $\delta F_{ext}$.

Since we assume the thruster magnitude is bounded with no impulsive behavior, it is not directly considered in obtaining the impact map. The swing foot end position is denoted by $p_2$ and $E_2 = \partial p_2 / \partial q_u$ denotes its Jacobian matrix, which yields the following relationship between the external forces and generalized forces $F_{ext} = E_2^\top F_2$, $F_2 = [F_2^x ; F_2^y]$. No rebound and no slippage are modeled by $E_2\dot{q}=0$. This gives $N+2$ equations from the integration of the unpinned dynamics during the impact, and $2$ equations from the constraints and $N+4$ unknowns for the states and external forces which can be written in the form of
\begin{equation}
    \begin{bmatrix}
        D_u && -E_2^\top \\ E_2 && 0
    \end{bmatrix} \begin{bmatrix}
        \dot{q}_u^+ \\ F_2
    \end{bmatrix} = \begin{bmatrix}
        D_u \\ 0
    \end{bmatrix}.
\end{equation}
Solving this system results in \eqref{eq:ImpF2} for computing the impact forces, where the superscript ($-$) indicates pre-impact values.
\begin{equation}
    \label{eq:ImpF2}
    F_2 = \Delta_{F_{2}}\dot{q}^-=-\left(E_{2} D_{u}^{-1} E_{2}^{\top}\right)^{-1} E_{2}\left[\begin{array}{c}
    {I_{N}} \\
    0
    \end{array}\right]\dot{q}^-
\end{equation}
\noindent After restricting the impact to the zero dynamics, writing the explicit dependency of each term results in
\begin{equation}
\begin{aligned}
    F_2 &= \Delta_{F_2}(q_f)\dot{q}_f \\
    &= \Delta_{F_2}(h_d(\alpha_f), \alpha_f)H^{-1}\begin{bmatrix}
        \frac{\partial h_d}{\partial \alpha}(\alpha_f) \\ 1
    \end{bmatrix}\kappa_1(\alpha_f)\sigma_{N}^-,
\end{aligned}
\end{equation}
\noindent which means $\sigma_N>0$ along the step $F_2^h = b_{F_2}^h(\alpha_f)\sigma_{N}^-$ and $F_2^v = b_{F_2}^v(\alpha_f)\sigma_{N}^-$. Let $\mu$ is the desired friction constant. Therefore, if for some $\zeta^->0$, $F_2^v>0$ and $(F_2^h/F_2^v)<\mu$ hold, then they also hold for all $\zeta^*>0$.
%
%
Furthermore, since the thrusters cannot produce impulsive signals, their values at the moment of impact do not affect the impact forces.

\subsection{Swing Phase Contact Force Contraints}

To compute the swing phase contact forces, consider the unpinned configuration variables $q_u = [q_b ; q_N ; x_1 ; y_1]$ and the unpinned dynamic model in \eqref{eq:UnpMod}, where $F_1 = [F_1^x ; F_1^y]$ are the contact forces on the stance foot along the $x$ and $y$ axis.
\begin{equation}
    \label{eq:UnpMod}
    %
        D_u\ddot{q}_u+\Omega_u=\underbrace{\begin{bmatrix}
            I_{N-1} & b_b \\ 0_{1 \times N-1} & b_N \\ 0_{2 \times N-1} & 0_{2 \times 2}
        \end{bmatrix}}_{B_1}\begin{bmatrix}
            u \\ F_T
        \end{bmatrix} + \begin{bmatrix}
            0_{N \times 2} \\ I
        \end{bmatrix}\underbrace{\begin{bmatrix}
            F_1^x+b_xF_T \\ F_1^y+b_yF_T
        \end{bmatrix}}_{F_r}
\end{equation}
\noindent We assume a stable supervisory controller enforces the virtual constraints $h_d(\alpha)$ in finite time. For any desired value of $b_NF_T$, the resulting force over the stance foot can be computed using the closed-loop unpinned system in
\begin{equation}
    \label{eq:CLSys}
    \dot{x}_u = \begin{bmatrix}
        \dot{q}_u \\ \ddot{q}_u
    \end{bmatrix} = \underbrace{\begin{bmatrix}
        \dot{q}_u \\ D_u^{-1}\left(B_1\begin{bmatrix}
            u \\ F_T
        \end{bmatrix}-\Omega_u\right)
    \end{bmatrix}}_{f_u}+\underbrace{\begin{bmatrix}
        0 \\ D_u^{-1}\begin{bmatrix}0 \\ I\end{bmatrix}
    \end{bmatrix}}_{g_u}F_r.
\end{equation}
%
%
\noindent From \eqref{eq:CLSys}, $F_r$ can be computed through the feedback linearization terms for an output function $y_u = [x_1, y_1]^\top$. This results in a system of the form
\begin{equation}
    \label{eq:F1FS}
    F_r = -L_{g_u}L_{f_u}y_u^{-1}L^2_{f_u}y_u,
\end{equation}
\noindent where the Lie derivatives can be computed as
\begin{equation}
    \begin{array}{l}
        L_{g_u}L_{f_u}y_u = \bar{D}_{22} \\
        L^2_{f_u}y_u = \left[\bar{D}_{21} \ \bar{D}_{22}\right]\left(B_1\begin{bmatrix}
        u(q) \\ F_T
        \end{bmatrix}-G_u-C_u\right)
    \end{array}
\end{equation}
%
%
\begin{equation}
    \bar{D} = \begin{bmatrix}
        \bar{D}_{11} & \bar{D}_{12} \\ \bar{D}_{21} & \bar{D}_{22}
    \end{bmatrix} = D_u^{-1}.
\end{equation}
\noindent Note that \eqref{eq:F1FS} is quadratic in $\dot{q}_u$ (since $C_u$ is quadratic in $\dot{q}_u$) and, therefore, linear in $\zeta$ when restricted to the zero dynamics, which simplifies to
\begin{subequations}
    \label{eq:F1ZD}
    \begin{equation}
        F_r = \Lambda_2(\alpha)F_T+\Lambda_1(\alpha)\zeta^*+\Lambda_0(\alpha),
    \end{equation}
    \begin{equation}
        F_r = \begin{bmatrix}\Lambda_2^h(\alpha) \\ \Lambda_2^v(\alpha)\end{bmatrix}F_T+\begin{bmatrix}\Lambda_1^h(\alpha) \\ \Lambda_1^v(\alpha)\end{bmatrix}\zeta^*+\begin{bmatrix}\Lambda_0^h(\alpha) \\ \Lambda_0^v(\alpha)\end{bmatrix}.
    \end{equation}
\end{subequations}
\noindent Functions $\Lambda_0(\alpha)$, $\Lambda_1(\alpha)$ and $\Lambda_2(\alpha)$ can be approximated by polynomial functions $L_0(\alpha) = [L_0^h(\alpha) ; L_0^v(\alpha)]$, $L_1(\alpha) = [L_1^h(\alpha) ; L_1^v(\alpha)]$ and $L_2(\alpha) = [L_2^h(\alpha) ; L_2^v(\alpha)]$ for a given nominal walking gait and then can be used as constraints for the online optimizer, as shown in \eqref{eq:F1CT}.
\begin{equation}
    \begin{array}{c}
        \label{eq:F1CT}
        \displaystyle\min_{\alpha_i\leq\alpha\leq\alpha_f}\left[(L_2^v(\alpha)-b_y)F_T+L_1^v(\alpha)\zeta^*+L_0^v(\alpha)\right] >0\\
        \displaystyle\max_{\alpha_i\leq\alpha\leq\alpha_f}\left[\frac{(L_2^h(\alpha)-b_x)F_T+L_1^h(\alpha)\zeta^*+L_0^h(\alpha)}{(L_2^v(\alpha)-b_y)F_T+L_1^v(\alpha)\zeta^*+L_0^v(\alpha)}-\mu\right] < 0
    \end{array}
\end{equation}
Note that the functions $L_0$, $L_1$ and $L_2$ can be pre-computed for a given nominal gait, allowing for those constraints to be enforced even on a online control framework that use the thrusters to improve on the nominal gait.

\section{Concluding Remarks}
In this work, we try to contribute to the design of legged robots with capabilities boosted through thruster-assisted locomotion. Our long-term goal is the development of robots capable of negotiating unstructured environments, including land and air, by leveraging legs and thrusters collaboratively. By doing this, we demonstrated that thrusters can be powerful tools and add to the feedback design flexibility by considering two scenarios of when the thusters are slow or fast in comparison to the internal dynamics. We explored thruster effects on the gait limit cycles and proposed new design paradigms based on shaping these center manifolds with strong foliations. In addition, unilateral contact force feasibility conditions were resolved in an optimal control scheme. If the dynamics of the thrusters are much quicker than the zero dynamics of the robot then the overall closed-loop system can be treated as a two-time-scale problem wherein the thruster parameter changes are applied to shape the limit cycles. 


\bibliographystyle{IEEEtran}
\bibliography{ref.bib,zotero-ref.bib}

\end{document}